\pdfoutput=1

\documentclass[11pt]{article}

\usepackage{acl}

\usepackage{times}
\usepackage{latexsym}
\usepackage{graphicx} 
\usepackage[T1]{fontenc}

\usepackage[utf8]{inputenc}

\usepackage{microtype}
\usepackage{amsmath}
\usepackage{booktabs}
\usepackage{multirow}
\usepackage{caption}
\usepackage{subcaption}
\usepackage{makecell}
\title{Multi-hop Commonsense Knowledge Injection Framework for\\ Zero-Shot Commonsense Question Answering}

\author{Xin Guan, Biwei Cao, Qingqing Gao, Zheng Yin, Bo Liu, Jiuxin Cao\thanks{\ \  Corresponding author.}\\
        Southeast University  \\ 
        \texttt{\{xin\_guan,caobiwei,qingqing\_gao,z.yin,bliu,jx.cao\}@seu.edu.cn} \\
        }

\begin{document}
\maketitle
\begin{abstract}
Commonsense question answering (QA) research requires machines to answer questions based on commonsense knowledge. However, this research requires expensive labor costs to annotate data as the basis of research, and models that rely on fine-tuning paradigms only apply to specific tasks, rather than learn a general commonsense reasoning ability. As a more robust method, zero-shot commonsense question answering shows a good prospect. The current zero-shot framework tries to convert triples in commonsense knowledge graphs (KGs) into QA-form samples as the pre-trained data source to incorporate commonsense knowledge into the model. However, this method ignores the multi-hop relationship in the KG, which is also an important central problem in commonsense reasoning. In this paper, we propose a novel multi-hop commonsense knowledge injection framework. Specifically, it explores multi-hop reasoning paradigm  in KGs that conform to linguistic logic, and we further propose two multi-hop QA generation methods based on KGs. Then, we utilize contrastive learning to pre-train the model with the synthetic QA dataset to inject multi-hop commonsense knowledge. Extensive experiments on five commonsense question answering benchmarks demonstrate that our framework achieves state-of-art performance.
\end{abstract}

\section{Introduction}
Commonsense knowledge is an essential basis for daily human communication and is also a key research area in the present natural language understanding system. In order to explore and improve the understanding and reasoning ability of machines for commonsense, many question answering (QA) benchmark datasets have been proposed, e.g.,  CommonsenseQA \cite{DBLP:conf/naacl/TalmorHLB19}, Abductive NLI \cite{DBLP:conf/iclr/BhagavatulaBMSH20}, PhysicalIQA \cite{DBLP:conf/aaai/BiskZLGC20}, etc. Meanwhile, with the development of large-scale pre-trained language models \cite{DBLP:conf/naacl/DevlinCLT19,DBLP:journals/corr/roberta,DBLP:journals/jmlr/RaffelSRLNMZLL20}, the gap between the performance of machines and humans in these datasets is gradually closing. However, some studies \cite{DBLP:journals/corr/Mitra,DBLP:conf/aaai/MaIFBNO21,DBLP:conf/naacl/KimKKAHY22} point out that models fine-tuned for specific downstream tasks are fitting individual datasets rather than learning a general commonsense reasoning ability. Therefore, as a more comprehensive evaluation method, zero-shot commonsense question answering has gradually become the focus of future research. 

\begin{figure}[t]
    \centering
    \includegraphics[width=0.99 \columnwidth]{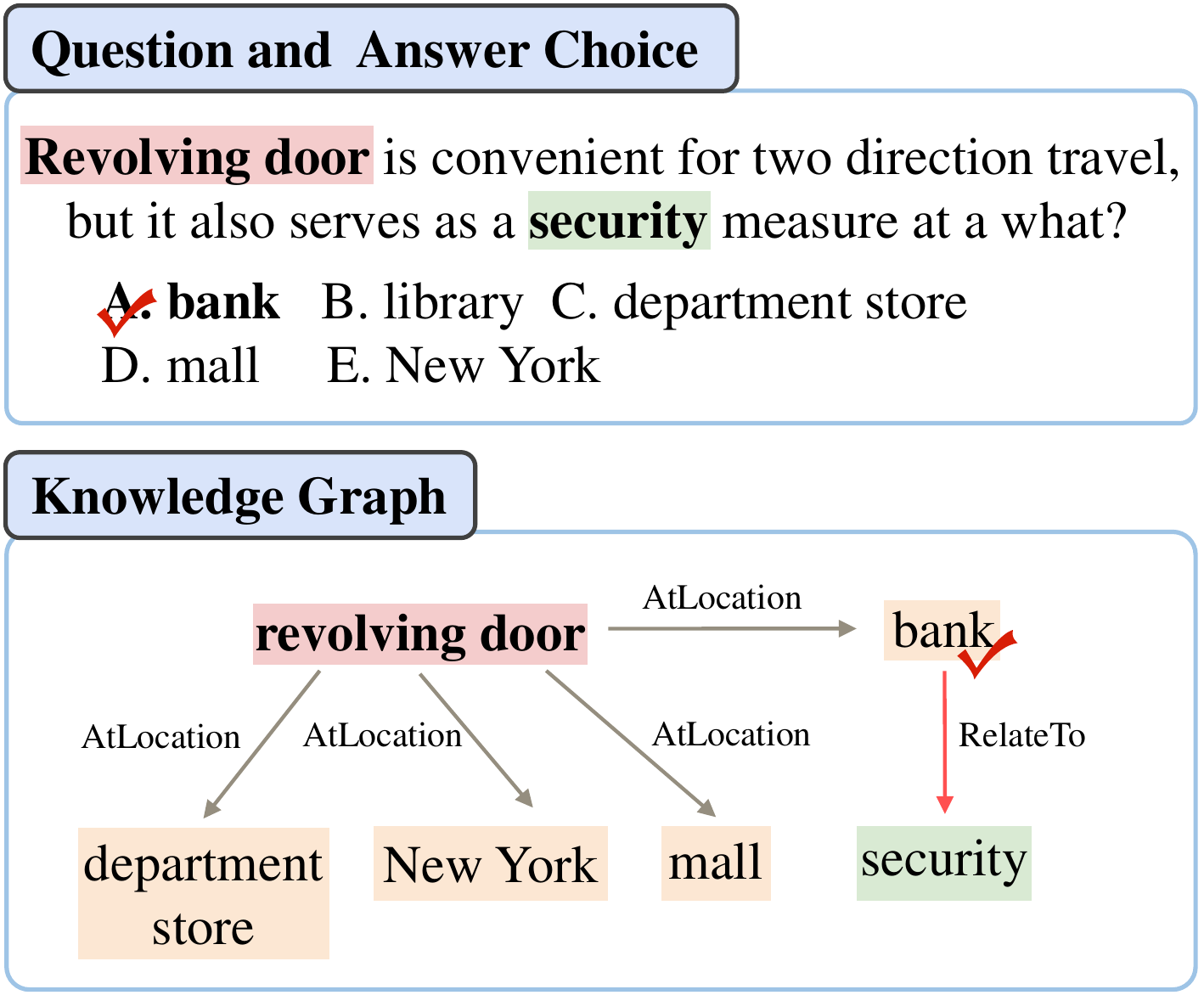} 
   \caption{Example of multi-hop commonsense reasoning. Two-hops path of knowledge is required to reason out the correct answer.}
    \label{fig1}
\end{figure}

Zero-shot commonsense question answering is to build a general commonsense question answering system with reasoning ability, rather than target task-specific scenarios. Recent work \cite{DBLP:conf/aaai/MaIFBNO21,DBLP:conf/naacl/KimKKAHY22} uses multi-source commonsense knowledge graphs, such as ConceptNet \cite{DBLP:conf/aaai/SpeerCH17}, Atomic \cite{DBLP:conf/aaai/SapBABLRRSC19}, and WordNet \cite{DBLP:journals/cacm/Miller95}, as data sources, and uses single-hop triples in knowledge graphs to generate synthetic QA datasets for training the QA model. However, most commonsense questions require multi-hop knowledge for correct reasoning. For these questions, the existing models do not have the ability of multi-hop reasoning, so they will choose the wrong answer.
As shown in Figure \ref{fig1}, for this commonsense question, it is necessary to reason the correct answer through the two-hop path of \emph{(revolving door, at location, bank)} and \emph{(bank, related to, security)} in the knowledge graph. However, previous work only generated QA samples based on single-hop triplets. The model trained by the dataset generated in this method does not have the reasoning ability for this multi-hop commonsense question. Since there are four answer entities directly connected with question entities in the knowledge graph, this model even misleads the answer selection by considering these four entities to be the correct answer.

To address the above problem, we propose a novel multi-hop commonsense knowledge injection framework for zero-shot commonsense question answering, which is divided into two steps: \emph{the synthetic multi-hop QA generation} and \emph{the knowledge injection pre-training}. 
 Specifically, in the process of synthetic multi-hop QA generation, we explore the general paradigm of multi-hop commonsense reasoning, and further propose two multi-hop QA generation methods with linguistic logic, compositive commonsense QA generation method and conjunctive commonsense QA generation method, which correspond to the continuous reasoning ability and multi-faceted reasoning ability of humans respectively. We generate a synthetic multi-hop QA dataset by these two methods.
 In the knowledge injection pre-training process, we utilize contrastive learning to pre-train the model on the synthetic multi-hop QA dataset to inject multi-hop commonsense knowledge.
 Finally, we can get a general commonsense question answering model with multi-hop knowledge reasoning ability.

The main contributions of this work are summarized as follows:
\begin{itemize}
    \item To our best knowledge, the proposed multi-hop commonsense knowledge injection framework is the first work to introduce multi-hop commonsense knowledge to improve the performance of zero-shot commonsense question answering.
\end{itemize}
\begin{itemize}
    \item We explore general paradigms of multi-hop commonsense reasoning in KG, and further propose two multi-hop QA generation methods with linguistic logic to generate a synthetic multi-hop QA dataset. Both methods can generate highly confusing negative samples to improve the performance of the model.
\end{itemize}
\begin{itemize}
    \item We utilize contrastive learning to pre-train on synthetic multi-hop QA datasets to get a commonsense question answering model with general reasoning ability and multi-hop knowledge.
\end{itemize}
\begin{itemize}
    \item We conduct extensive experiments on various commonsense question answering benchmarks. The results demonstrate that our framework achieves state-of-art performance.
\end{itemize}

\section{Related Work}
\subsection{Zero-Shot Commonsense Question answering}
Zero-shot commonsense question answering focuses on constructing unsupervised models without any label supervision. Current work can be divided into two paradigms: (1) Designing unsupervised models by leveraging the properties of pre-trained language models. Some work utilizes the masked language model (MLM) task in autoregressive language models, such as Word2Vec, BERT \cite{DBLP:conf/naacl/DevlinCLT19}, to evaluate the plausibility of sentences composed of question and answer \cite{DBLP:conf/acl/TamborrinoPPVN20,DBLP:journals/corr/abs-1806-02847}. Other works take generative language models, such as GPT-2 \cite{radford2019language} and GPT-3 \cite{DBLP:conf/nips/BrownMRSKDNSSAA20}, to design template prefixes to generate commonsense knowledge or answers \cite{DBLP:conf/emnlp/ShwartzWBBC20,DBLP:conf/acl/NiuHLC0H20,DBLP:conf/acl/0010LLWWBCH22,wang-zhao-2022-art}. (2) Injecting knowledge into pre-trained models. These works utilize external knowledge (e.g., ConceptNet, ATOMIC, Wiki) as data sources and design different pre-training tasks to incorporate knowledge into the pre-trained models \cite{DBLP:conf/emnlp/BanerjeeB20,DBLP:conf/aaai/MaIFBNO21,DBLP:conf/naacl/KimKKAHY22}. Since the injected knowledge enables the model to learn relevant commonsense knowledge through pre-training, this method is more competitive on some commonsense question answering tasks. Our work also focuses on this zero-shot method. 

\begin{figure*}[t]
    \centering
    \includegraphics[width=0.99 \textwidth]{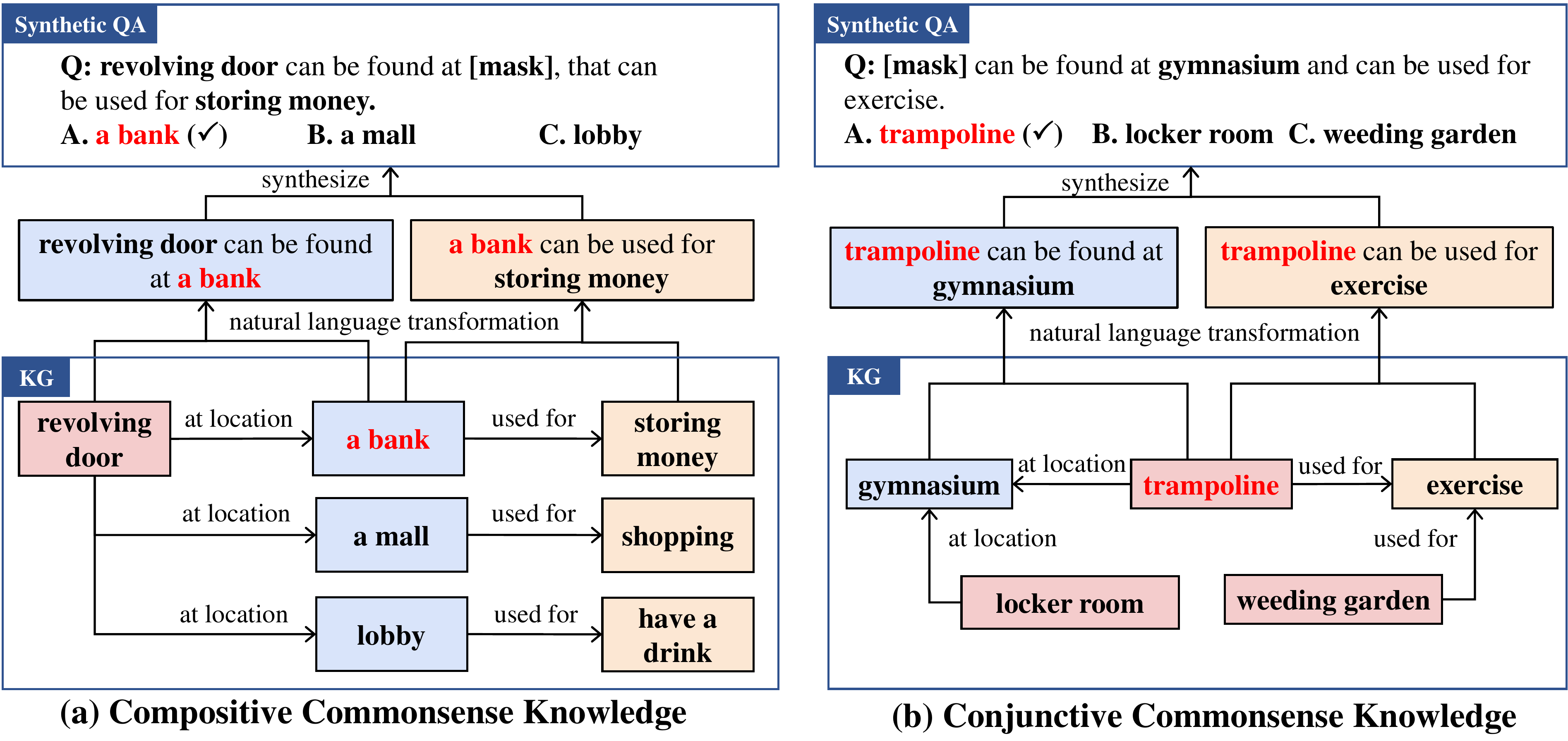} 
   \caption{Synthetic multi-hop QA generation process. \textbf{(a)} is the compositive commonsense QA generation process and \textbf{(b)} is the conjuctive commonsense QA generation process. Each QA generation method extracts multi-hop knowledge containing positive and negative samples in KG, and converts them into QA-form samples according to natural language templates.} 
    \label{fig2}
\end{figure*}

\subsection{Commonsense Question Answering with External Knowledge}
In the supervised setting, there have been many work attempts to incorporate external knowledge reasoning modules into the model. These works \cite{DBLP:conf/emnlp/LinCCR19,DBLP:conf/emnlp/FengCLWYR20,DBLP:conf/naacl/YasunagaRBLL21,DBLP:conf/coling/GuanCGY0C22} extract subgraphs related to QA concepts in the knowledge graph, and design graph encoders, combined with text encoders for reasoning. In the unsupervised scene, many works directly use the knowledge graph as a data source for pre-training. \citet{DBLP:conf/emnlp/BanerjeeB20} design the knowledge triplet learning, using any two elements to predict the remaining element. \citet{DBLP:conf/aaai/MaIFBNO21} construct question answering samples using knowledge graphs, and train the model with the masked language modeling method. \citep{DBLP:conf/naacl/KimKKAHY22} design knowledge adapters to mitigate the loss of knowledge from the interference among the different knowledge sources. However, the above unsupervised methods only use single-hop triples as training samples, ignoring the rich multi-hop relationships in the knowledge graph. We propose a multi-hop knowledge injection framework to solve this problem.

\section{Multi-hop Commonsense Knowledge Injection Framework}
In this paper, we focus on the zero-shot commonsense knowledge question answering task, where the model or system can not access any training or validation data for commonsense tasks. Under this setting, we first generate a synthetic QA dataset from KG, and then pre-train the model on this dataset to inject commonsense knowledge. Finally, we test our model on different commonsense question answering tasks to verify the model's commonsense reasoning ability.
 
In previous zero-shot commonsense question answering frameworks \cite{DBLP:conf/aaai/MaIFBNO21,DBLP:conf/naacl/KimKKAHY22}, four knowledge graphs, \emph{ConceptNet} \cite{DBLP:conf/aaai/SpeerCH17}, \emph{WordNet} \cite{DBLP:journals/cacm/Miller95}, \emph{WikiData} \cite{DBLP:journals/cacm/VrandecicK14}, \emph{ATOMIC} \cite{DBLP:conf/aaai/SapBABLRRSC19}, are used as data sources to synthesize QA. Formally, for a triplet $(e^{head}, r, e^{tail})$ in KG, where $e^{head}, r, e^{tail}$ denote head entity, relation and tail entity respectively, previous method converts $e^{head}$ and $r$ into question $Q$ through natural language templates, takes $e^{tail}$ as the correct answer $A_c$, and randomly selects tail entities in other triplets as distractors $\{A_1, \cdots, A_n\}$, where $n$ is the number of negative samples. By above method, one synthesized QA sample $(Q,A)$ is generated, where $A=\{A_c, A_1, \cdots, A_n\}$. After generating the synthetic dataset, it uses the model to perform a masked language modeling task on this dataset to obtain commonsense knowledge.

Our work builds on previous frameworks, addressing the lack of multi-hop reasoning capabilities in current zero-shot commonsense question answering research.

\subsection{Synthetic Multi-hop QA Generation}
We follow the definition in the previous frameworks and propose two methods for synthetic multi-hop QA generation on this basis, as shown in Figure \ref{fig2}. During the generation process, we follow two criteria: (1) The synthetic multi-hop QA needs to have practical significance, conforming to daily dialogue scenarios and human logical thinking. (2) The distractors should have a strong interference ability to improve the reasoning performance of the model. It needs to have a semantic relationship with the question while also clearly distinguished from the correct answer. The following are the proposed generation methods based on these criteria.

\paragraph{Compositive Commonsense QA Generation.} Compositive commonsense QA is generated by triples with the logical relationship in the knowledge graph. This is the most common form of multi-hop commonsense knowledge, which shows the continuous reasoning ability of humans. For example, when you see the word ``university'', you will think of ``library'', and then think of ``books''. Our proposed generation method is as follows, divided into two steps. \textbf{(1) Generating question and answer.} Formally, given a triple $(e_1^{head}, r_1, e_1^{tail})$, we find another triple $(e_2^{head}, r_2, e_2^{tail})$, while $e_1^{tail}=e_2^{head}=e^{key}$. Each triple will be converted into a sentence through the natural language template. Then, we integrate these two sentences and mask the connected entities $e^{key}$ in them as the question $Q$. Meanwhile, entity $e^{key}$ is also the correct answer $A_c$.
\textbf{(2) Generating hard negative samples (distractors).} We generate distractors following the above criteria as follows. Formally, for each distractor entity $e_3$, in order to ensure it is relevant to the question, it needs to satisfy the condition: $(e_1^{head}, r_1, e_3)$. Meanwhile, to ensure that it is an error option, it needs to satisfy the condition: a): $e_3 \ne e^{key}$, b): for any $(e_3, r_2, e_3^{tail}), e_3^{tail}\ne e_2^{tail}$. We take $e_3$ as a distractor $A_i$ and select the distractor set $\{A_1, \cdots, A_n\}$ in this way. Finally, we can get a compositive commonsense QA pairs $(Q, A)$ sample. Figure \ref{fig2} (a) shows an example of this process.

\begin{figure*}[t]
    \centering
    \includegraphics[width=0.99 \textwidth]{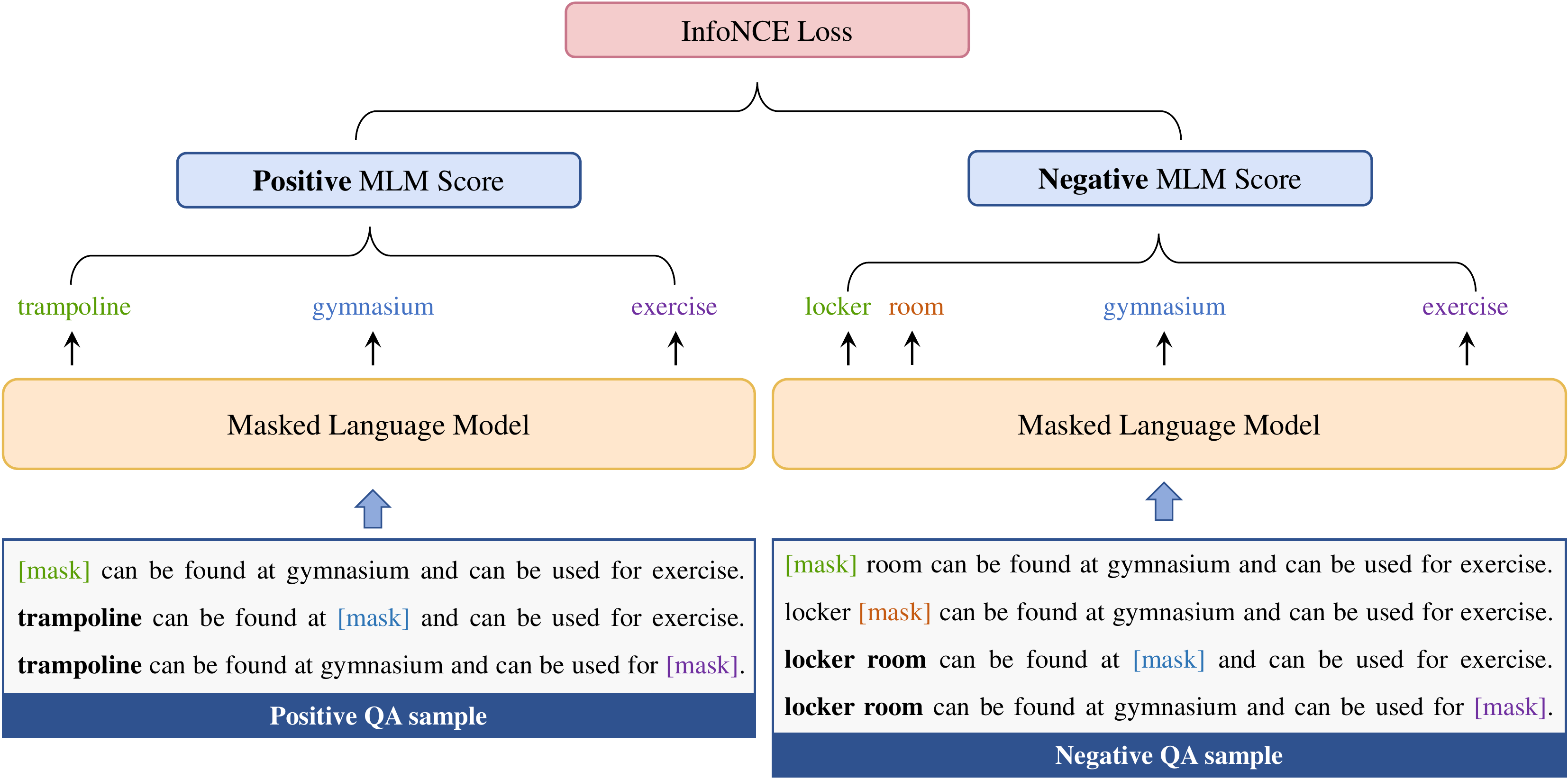} 
   \caption{The overview of Knowledge injection pre-training process. We concatenate question with each answer option to build positive and negative samples. Then we mask the tokens one by one and compute the MLM loss as the score of samples. Finally, we use InfoNCE loss from contrastive learning to optimize the model. }
    \label{fig3}
\end{figure*}

\paragraph{Conjunctive Commonsense QA Generation.} Conjunctive commonsense QA is generated by entities with multiple attributes in the knowledge graph, which shows that human thinking ability is multi-faceted. For example, when you see a ``gym'', you think of ``basketball'', ``football'', and other sports simultaneously. We also generate QA in two steps. \textbf{(1) Generating question and answer.} Formally, given a triple $(e_1^{head}, r_1, e_1^{tail})$, we find another triple $(e_1^{head}, r_2, e_2^{tail})$, while $e_1^{tail} \ne  e_2^{tail}$ and $e_1^{head}=e^{key}$. Each triple will be converted into a sentence through the natural language template. Then, we integrate these two sentences and mask the connected entities $e^{key}$ in them as the question $Q$. Meanwhile, entity $e^{key}$ is also the correct answer $A_c$.
\textbf{(2) Generating hard negative samples (distractors).} Formally, for each distractor entity $e_3$, in order to ensure it is relevant to the question, it needs to satisfy one of the following conditions: a) $(e_3, r_1, e_1^{tail})$, b) $(e_3, r_2, e_2^{tail})$. Meanwhile, to ensure that it is an error option, for the above two conditions, only one of them can be satisfied and the other can not be satisfied. We take $e_3$ as a distractor $A_i$ and select the distractor set $\{A_1, \cdots, A_n\}$ in this way. Finally, we can get a conjunctive commonsense QA pairs $(Q, A)$ sample. Figure \ref{fig2} (b) shows an example of this process.

By the above method, we generate the synthetic multi-hop QA dataset and merge it with the single-hop QA dataset from the previous work \cite{DBLP:conf/aaai/MaIFBNO21} for the subsequent pre-training task.

\subsection{Knowledge Injection Pre-training}
The overview of our knowledge injection pre-training method is shown in Figure \ref{fig3}. We use the synthetic multi-hop QA dataset for pre-training to inject multi-hop commonsense knowledge. Specifically, we use the masked language model, RoBERTa \cite{DBLP:journals/corr/roberta}, for pre-training. Given a synthesized QA pairs $(Q,A)$ sample, where $A=\{A_c, A_1, \cdots, A_n\}$. We concatenate question $Q$ with each answer option $A_i$ to build the the input QA sequence $T = \{T^P_0, T^N_1, \cdots ,T^N_n\}$, where $T^P_0 = [Q;A_c]$ is the positive QA sample and $T^N_i = [Q;A_i]$ is the negative QA sample. For each input sample, we mask each token one by one and compute the sum of their MLM losses as the score $S$ as following:
\begin{equation} \label{eq1}
S(T_i)=-\frac{1}{m} \sum_{j=1}^m \log P\left( \ldots t_{j-1}, t_{j+1}, \ldots; \theta \right)
\end{equation}
where $t_{j}$ is a word token in the input QA sequence $T_i$, $m$ is the length of $T_i$, $P$ is the conditional probability from masked language model parameterized by $\theta$.

After getting the scores of all samples, our purpose is to make the positive sample have the lowest score and make the negative sample score higher than that of the positive sample. Here we use the idea of contrastive learning to optimize the model. To this end, we invert all values and use Softmax for normalization. In this way, the positive sample should have a score of 1, and the negative sampless have a score of 0. We use InfoNCE loss \cite{DBLP:journals/corr/abs-1807-03748}, a form of a contrastive loss function, to optimize the model as following:
\begin{equation}
\mathcal{L}=-\log \frac{\exp \left(-S(T^P) / \tau\right)}{\sum_{i=0}^n \exp \left(-S(T_i) / \tau\right)}
\end{equation}
where $\tau$ is a temperature hyper-parameter. The sum is over one positive QA sample and $n$ negative QA samples.

In the evaluation of the model, we construct the input QA samples in the same way, and calculate the score of each answer option according to Equation \ref{eq1}, and finally choose the correct answer with the lowest score.

\section{Experiments}
In this section, We evaluate our framework on five commonsense question answering benchmarks and conduct ablation experiments to verify the effectiveness of each part of our framework. 

\subsection{Experimental Settings}
Our framework is under the zero-shot setting, that is, the model cannot have access to any official training data of any benchmarks during the training process. For the evaluation, we use the validation set of each benchmark (some test sets are not publicly available). Since the validation set is not used for hyperparameter tuning here, it can be regarded as the test set.

\subsection{Benckmarks}
We evaluate our proposed framework on five question-answering benchmarks for commonsense reasoning.
\paragraph{CommonsenseQA (CSQA)} \cite{DBLP:conf/naacl/TalmorHLB19}: It is a multiple choice question QA dataset that requires conceptual commonsense knowledge for reasoning. Questions and answers are artificially generated according to entities in ConceptNet and their relations.
\paragraph{Multiple Choice Temporal commonsense (MC-TACO)} \cite{DBLP:conf/emnlp/ZhouKNR19}: This dataset requires temporal commonsense comprehension, such as duration, temporal ordering, typical time, frequency, and stationarity.
\paragraph{Abductive NLI (a-NLI)} \cite{DBLP:conf/iclr/BhagavatulaBMSH20}: Abductive reasoning is inference to the most plausible explanation. Each sample is a real event, which has two potential explanations for the given situation. 
\paragraph{PhysicalIQA (PIQA)} \cite{DBLP:conf/aaai/BiskZLGC20}: This is a dataset about physical commonsense knowledge (e.g., a bucket can hold paint). It needs to select the solution for the given goal among two choices.
\paragraph{Wino-Grande (WG)} \cite{DBLP:journals/cacm/SakaguchiBBC21}. It is a dataset that is crowdsourced with a carefully designed procedure to improve the scale and robustness against the dataset-specific bias. It is formulated as a fill-in-a-blank task with binary options.
\begin{table*}
\centering
\resizebox{\textwidth}{!}{
\begin{tabular}{lcccccc}
\toprule 
\textbf{Model} & \textbf{KG} & \textbf{CSQA} & \textbf{MC-TACO} & \textbf{a-NLI} & \textbf{PIQA} & \textbf{WG} \\
\midrule 
Random & - & 20.0 & - & 50.0 & 50.0 & 50.0 \\
Majority & - & 20.9 & - & 50.8 & 50.5 & 50.4 \\
GPT2-Large  & - & 41.4 & 64.5 & 56.5 & 68.9 & 53.2 \\
RoBERTa-Large & - & 45.0 & 57.7 & 65.5 & 67.6 & 57.5 \\
Self-talk \cite{DBLP:conf/emnlp/ShwartzWBBC20} & - & 32.4 & 59.9 & - & 70.2 & 54.7 \\
SMLM \cite{DBLP:conf/emnlp/BanerjeeB20} & * & 38.8 & - & 65.3 & - & - \\
RoBERTa-L (MR) \cite{DBLP:conf/aaai/MaIFBNO21} & AT & 64.2 & - & - & 72.1 & 59.2 \\
RoBERTa-L (MR) \cite{DBLP:conf/aaai/MaIFBNO21} & CN,WD,WN & - & - & 70.0 & 72.0 & 59.4 \\
RoBERTa-L (MR) \cite{DBLP:conf/aaai/MaIFBNO21} & Whole & 67.4 & 65.6 & 70.5 & 72.4 & 60.9 \\
Zero-shot fusion (Adapter) \cite{DBLP:conf/naacl/KimKKAHY22} & Whole & 68.2 & - & 72.5 & 72.9 & 60.8 \\
\midrule 
\textbf{Multi-hop Knowledge Inject Framework} & Whole & \textbf{71.0}($\pm$0.3) & \textbf{70.9}($\pm$1.0) & \textbf{72.5}($\pm$0.3) & \textbf{73.1}($\pm$0.3) & \textbf{61.0}($\pm$0.3) \\
\bottomrule 
\end{tabular}
}
\caption{Zero-shot evaluation results with different combinations of models and knowledge sources across five commonsense tasks. The best results of each benchmark are in bold. AT, CN, WD and WN represent ATOMIC, ConceptNet, WikiData and WordNet, respectively. Whole represents the combination of AT, CN, WD and WN. SMLM (*) used different KGs for the different benchmark.}
\label{tab1}
\end{table*}

\subsection{Knowledge Graphs}
In our framework, we use four KGs: ATOMIC \cite{DBLP:conf/emnlp/SapRCBC19}, ConceptNet \cite{DBLP:conf/aaai/SpeerCH17}, WikiData \cite{DBLP:journals/cacm/VrandecicK14} and WordNet \cite{DBLP:journals/cacm/Miller95}. Each KG has different types of knowledge. ATOMIC focuses on social commonsense knowledge, such as the cause or follow-up of an event. ConceptNet contains general conceptual commonsense knowledge, describing the relationship between a conceptual entity and another conceptual entity. WikiData is a general KG collected by Wikipedia. WordNet is a lexical database of semantic relations between words. In the synthetic single-hop QA generation, we follow the work of \citet{DBLP:conf/aaai/MaIFBNO21} and use the above four KGs for generation. In our synthetic multi-hop QA generation, we choose ConceptNet as the data source for generation, because ConceptNet has an obvious multi-hop relationship compared with other KGs.

\subsection{Implementation}
Our framework is implemented with PyTorch and RoBERTa-Large \cite{DBLP:journals/corr/roberta} from the Hugging face Transformers library. In our experiments, we use max sequence length 128, warm-up proportion 0.05, weight decay 0.01, adam $\beta_1$ 0.9, adam $\beta_2$ 0.98, adam epsilon $1e^{-6}$, batch size 2 per GPU, for only 1 epoch in the synthetic multi-hop QA dataset. The training process is conducted on 4 * NVIDIA Tesla T4 (15G) and costs about 12 hours in total. We run our experiments with different random seeds. 

In the synthetic multi-hop QA generation process, we generate two negative samples for each QA. In the knowledge injection pre-training process, we randomly sample 95\% of the synthetic QA dataset for training while the remaining 5\% are used for validation. The temperature $\tau$ in InfoNCE loss is set to 0.7. 

\subsection{Baselines}
We compare our framework with the following baselines.  \textbf{Random} method is to randomly take the label as the answer and \textbf{Majority} method is to take the most frequent label as the answer. We take the pre-trained language model, \textbf{RoBERTa-Large} \cite{DBLP:journals/corr/roberta} and \textbf{GPT2-Large} \cite{radford2019language} without fine-tuning, for comparison.

\paragraph{Self-talk} \cite{DBLP:conf/emnlp/ShwartzWBBC20} combines context and pre-defined template prefixes, which are used LM to generate clarification prompts, and then elicit knowledge from another LM. Finally, it evaluates each answer candidate with the original text and generated knowledge.

\paragraph{SMLM} \cite{DBLP:conf/emnlp/BanerjeeB20} is a method based on knowledge triplet learning, which masks any one of the triplets and use the remaining two for inference.

\paragraph{RoBERTa-L(MR)} \cite{DBLP:conf/aaai/MaIFBNO21} uses different KG to generate the synthetic single-hop QA dataset and uses marginal ranking loss to train the RoBERTa model. 

\paragraph{Zero-shot fusion (Adapter)} \cite{DBLP:conf/naacl/KimKKAHY22} uses different KG to generate the synthetic single-hop QA dataset separately. It introduces an expert adapter for each dataset for training, and finally fuses these expert adapters for evaluation.

\subsection{Main Results}
Table \ref{tab1} shows the zero-shot evaluation results on five commonsense tasks. Our framework achieves the best performance across all baseline models. In particular, our framework greatly improves the accuracy by \textasciitilde2.8\% and \textasciitilde5.3\% in the CSQA and MC-TACO benchmarks respectively. 

It is worth noting that though our framework has a slight improvement on some benchmarks compared to zero-shot fusion (Adapter) \cite{DBLP:conf/naacl/KimKKAHY22}, they introduce expert adapters for each KG separately, which increase the number of parameters of the model. And they train more epochs on each separate synthetic dataset, using more synthetic data (since they didn't filter the data). This means their method is not friendly used in the low-resource computing setting. RoBERTa-L (MR) \cite{DBLP:conf/aaai/MaIFBNO21}, as a method with the same model architecture as ours, comparing with it can better reflect the effectiveness of our framework. It can be clearly observed that our framework is significantly improved compared to RoBERTa-L (MR) on multiple benchmarks. This also reflects that our proposed multi-hop knowledge injection framework can enable the model to learn multi-hop knowledge.

\begin{table}
    \centering
    \resizebox{0.9\columnwidth}{!}{
    \begin{tabular}{lc}
    \toprule
    \textbf{Methods} & \textbf{CSQA}  \\
    \midrule 
    Multi-hop Knowledge Inject Framework & 71.0 \\
    \midrule 
    \,w/o hard negative answer & 69.7   \\
    \,w/o compositive QA & 70.2   \\
    \,w/o conjunctive QA & 70.6   \\
    \,w/o multi-hop knowledge & 68.4   \\
    \midrule 
    \,w/o InfoNCE loss; +margin loss & 70.2\\
    \bottomrule 
    \end{tabular}
    }
    \caption{Ablation study results. We report the accuracy on CommonsenseQA. The middle rows are ablation experiments for synthetic multi-hop QA generation and the bottom row is the ablation experiments for knowledge injection pre-training. }
    \label{tab2}
\end{table}

\subsection{Ablation Study}

\paragraph{Synthetic Multi-hop QA Generation.}  To explore the impact of the proposed synthetic multi-hop QA dataset, we conduct multiple ablation experiments. 1) w/o hard negative answer: remove hard negative samples and replace them with ordinary negative samples (randomly select entities as negative samples). 2) w/o compositive QA: remove compositive commonsense QA generation method. 3) w/o conjunctive QA: remove conjunctive commonsense QA generation method. 4). w/o multi-hop knowledge: remove all multi-hop commonsense knowledge. Table \ref{tab2} shows the ablation study results. We find that removing the hard negative answers will significantly degrade performance by 1.3\%, which proves that our negative sample selection method is effective. Removing the compositive QA generation method or the conjunctive QA generation method will reduce the accuracy, demonstrating that the QA samples generated by these two methods have different aspects of multi-hop knowledge. Removing all multi-hop knowledge will significantly degrade the performance, which shows that our framework overcomes the shortcomings of QA methods generated from single-hop triplet data.

\begin{figure}[t]
    \centering
    \includegraphics[width=0.99 \columnwidth]{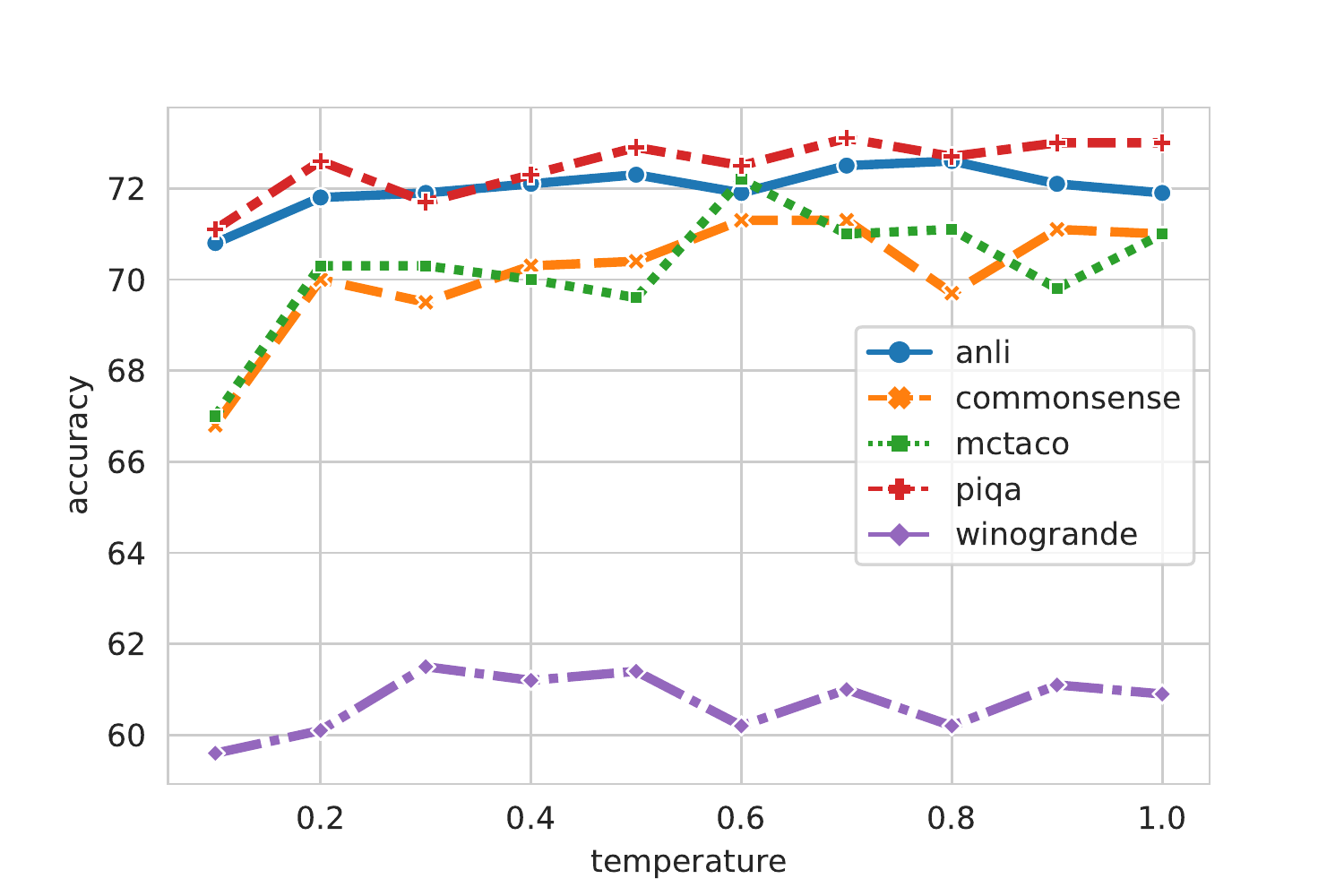} 
   \caption{Result with different temperatures across five commonsense tasks.}
    \label{fig4}
\end{figure}

\paragraph{Knowledge Injection Pre-training.} We conduct ablation experiments on the knowledge injection pre-training method. We remove InfoNCE loss and use margin loss for evaluation. Table \ref{tab2} shows that the performance of the model drops slightly, which also proves the effectiveness of our framework. Meanwhile, we train the model with different temperatures $\tau$ and evaluate them on the benchmarks as shown in Figure \ref{fig4}. The temperature $\tau$ corresponding to the best results of most datasets is between 0.6 and 0.8, except for the WG. We analyze that due to the small number of negative samples we select (two negative samples), our temperature $\tau$ is relatively high. Since the WG dataset involves more social commonsense knowledge and less conceptual commonsense knowledge, it is not sensitive to multi-hop commonsense knowledge and negative samples.

\subsection{Case Study}
\begin{table*}[t]
    \centering
    \includegraphics[width=0.99 \textwidth]{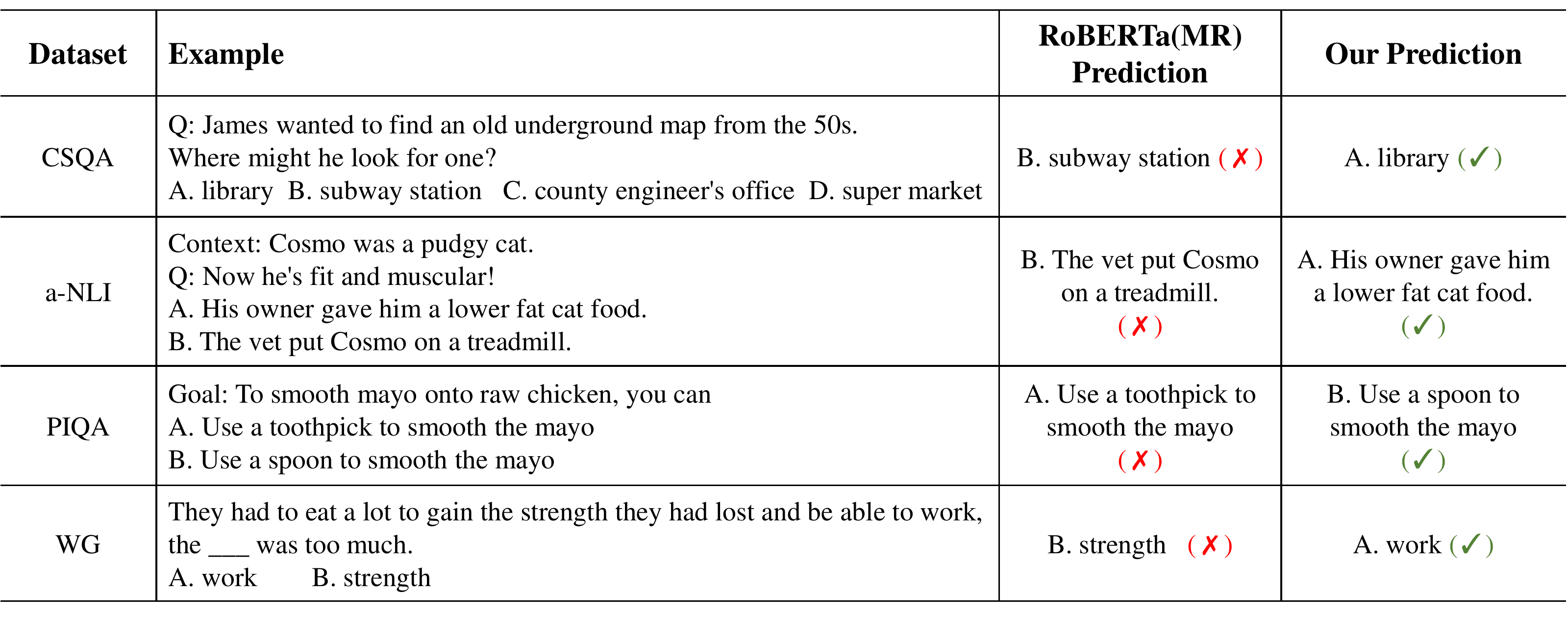} 
   \caption{Cases from different datasets. Compared with Roberta(MR), our framework can obtain more knowledge and perform better on different datasets.}
    \label{tab3}
\end{table*}

Table \ref{tab3} shows some cases from different datasets, where we compare predictions of our framework with the RoBERTa (LM) \cite{DBLP:conf/aaai/MaIFBNO21}. We observe that RoBERTa (LM) makes incorrect choices on some options that are highly confusing. For example, in the CSQA dataset, for the question ``James wanted to find an old underground map from the 50s. Where might he look for one?''  RoBERTa (LM) only captures the relationship between ``underground'' and ``subway station'', so it chooses the wrong answer. Our model captures the relationship between ``old'', ``underground map'', and ``library'', so it chooses the right answer. We also show other examples in the table to illustrate the effectiveness of our framework. These cases show that our framework can overcome the lack of multi-hop knowledge of previous models by injecting the multi-hop knowledge generated by our synthetic Multi-hop QA dataset.

\section{Conclusion}
In this paper, we propose a novel multi-hop commonsense knowledge injection framework for zero-shot commonsense question answering task to address the lack of multi-hop knowledge reasoning ability in the current zero-shot framework. Specially, this framework is divided into two steps: synthetic multi-hop QA generation and knowledge injection pre-training. In the process of QA synthesis, we explore the linguistically logical multi-hop reasoning patterns existing in KGs, and propose two QA generation methods based on KGs. Then, in the process of knowledge injection pre-training, we use the synthetic multi-hop QA dataset for pre-training with contrastive learning to inject multi-hop knowledge. We conduct extensive experiments on five commonsense question answering benchmarks, the results show that our framework achieves state-of-the-art performance. Ablation experiments and case study also confirm the effectiveness of our knowledge injection framework.

\section*{Limitations}
Although we have explored the multi-hop commonsense knowledge injection framework, we utilize one KG, ConceptNet. Since there is no explicit multi-hop relationship for other KGs, such as Atomic, we do not apply them. Therefore, exploring the implicit multi-hop relations in other KGs can be our future work. In addition, our synthetic multi-hop QA generation method applies entities and their relationships within two hops of an entity node as the center, and does not consider the rest of the nodes beyond two hops. For these nodes, we can consider designing new methods to utilize them to enrich more multi-hop knowledge.

\section*{Ethics Statement}
Commonsense question answering is an important field in question answering. 
The datasets and knowledge graphs used in our work are all public data, and the models we use are also public models in the field.
Since our framework uses data from existing datasets and KGs, it may also inherit the social biases present in these underlying datasets.
Our work is conform to to ACL Code of Ethics.

\bibliography{acl_latex}
\bibliographystyle{acl_natbib}




\end{document}